\documentclass[letterpaper, 10 pt, conference]{ieeeconf}  

\newcommand{\forarxiv}{1}

\IEEEoverridecommandlockouts                              

\usepackage{lipsum}
\usepackage[utf8]{inputenc}
\usepackage{amsmath}
\usepackage{graphicx}
\usepackage{grffile}
\usepackage{dsfont}
\usepackage{amsthm}
\usepackage[hidelinks]{hyperref}
\usepackage[capitalize]{cleveref}
\usepackage{eufrak}
\usepackage{subcaption}
\usepackage[linesnumbered,ruled,vlined]{algorithm2e}
\usepackage{amssymb}

\usepackage{float}
\newtheorem{theorem}{Theorem}[section]

\newtheorem{lemma}[theorem]{Lemma}

\usepackage{tikz}
\usetikzlibrary{shapes,arrows,chains}
\usepackage{wrapfig}
\crefname{section}{Section}{Sections}
\crefname{theorem}{Theorem}{Theorems}
\crefname{lemma}{Lemma}{Lemmas}
\crefname{table}{Table}{Tables}
\crefformat{equation}{(#2#1#3)}
\usepackage[sort,compress]{cite}
\usepackage{accents}

\DeclareMathOperator*{\argmax}{\arg\!\max}

\usepackage{array,multirow,graphicx}
\usepackage{colortbl}
\usepackage[detect-weight=true, detect-family=true]{siunitx}

\usepackage{mathtools}

\newcommand{\STAB}[1]{\begin{tabular}{@{}c@{}}#1\end{tabular}}

\if \forarxiv 1
\usepackage[final]{changes} 
\else
\usepackage[authormarkuptext=name,authormarkupposition=left]{changes}
\fi

\definechangesauthor[name={MFE}, color={orange}]{MFE}
\definechangesauthor[name={GH}, color={brown}]{GH}
\definechangesauthor[name={JH}, color={red}]{JH}

\title{\LARGE \bf
Robustness Analysis of Neural Networks via Efficient Partitioning \added{with} Applications in Control Systems
}

\author{Michael Everett,  Golnaz Habibi, Jonathan P. How
\thanks{The authors are with the Aerospace Controls Laboratory at the Massachusetts Institute of Technology, {\tt\small{\{mfe, ghabibi, jhow\}@mit.edu}}. This work was supported by Ford Motor Company.}%
}

\begin{document}

\maketitle
\thispagestyle{empty}
\pagestyle{empty}

\begin{abstract}

Neural networks (NNs) are now routinely implemented on systems that must operate in uncertain environments, but the tools for formally analyzing how this uncertainty propagates to NN outputs are not yet commonplace.
Computing tight bounds on NN output sets (given an input set) provides a measure of confidence associated with the NN decisions and is essential to deploy NNs on safety-critical systems.
Recent works approximate the propagation of sets through nonlinear activations or partition the uncertainty set to provide a guaranteed outer bound on the set of possible NN outputs.
However, the bound looseness causes excessive conservatism and/or the computation is too slow for online analysis.
This paper unifies propagation and partition approaches to provide a family of robustness analysis algorithms that give tighter bounds than existing works for the same amount of computation time (or reduced computational effort for a desired accuracy level).
Moreover, we provide new partitioning techniques that are aware of their current bound estimates and desired boundary shape (e.g., lower bounds, weighted $\ell_\infty$-ball, convex hull), leading to further improvements in the computation-tightness tradeoff.
The paper demonstrates the tighter bounds and reduced conservatism of the proposed robustness analysis framework with examples from model-free RL and forward kinematics learning.
\end{abstract}



\section{Introduction}
\label{sec:intro}
Neural networks (NNs) are ubiquitous across robotics for perception, planning, and control tasks.
While empirical performance statistics can indicate that a NN has learned a useful input-output mapping, there are still concerns about how much confidence to associate with decisions resulting from a learned system.
One direction toward providing a confidence measure is to consider how the various sources of uncertainty in training/execution processes map to uncertainty in outputs of trained NNs.
Many of these uncertainties appear at the NN input (e.g., from noisy/adversarially attacked sensing, unknown initial conditions), thus this work focuses on the problem of propagating input uncertainties through NNs to bound the set of possible NN outputs online.

Analysis of how a set of possible inputs propagates through a NN has an inherent tradeoff between computation time and conservatism.
Exact methods~\cite{Ehlers_2017, Katz_2017, Huang_2017b,Lomuscio_2017, Tjeng_2019,Gehr_2018} are computationally intractable for online analysis, so we focus on finding guaranteed outer bounds on the network outputs.
Most existing methods propagate the entire input set through the NN -- we refer to these as \textit{Propagators}~\cite{gowal2018effectiveness,Raghunathan_2018,fazlyab2019safety,zhang2018efficient,Weng_2018,singh2018fast}.

Although some of these propagators scale to high dimensional NNs, large input sets (e.g., from high state uncertainty) induce massive conservatism, even for small NNs.
\textit{Partitioners}~\cite{anderson2020tightened,xiang2018output,wang2018formal,rubies2019fast,xiang2020reachable} are a promising direction toward propagating large input sets through NNs, particularly when the number of uncertain NN inputs is relatively small, as in many control systems.
Nonetheless, current partitioners spend excessive computational effort when refining cells and suffer from simple propagation strategies.



\begin{figure}[t]
	\centering
	\includegraphics[page=6, width=0.9\linewidth, trim=0 120 0 60, clip]{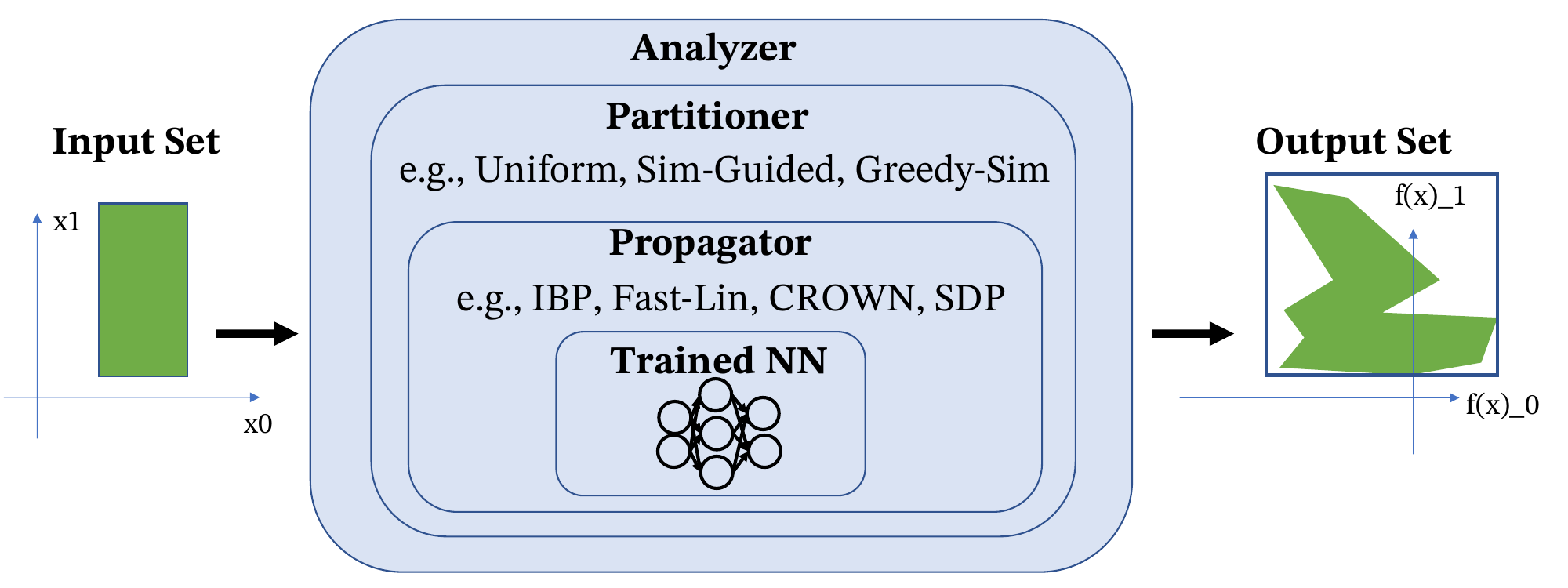}
	\vspace*{-.1in}
	\caption{
		Robustness Analysis Architecture. This work efficiently bounds the set of NN outputs for a given input set.
	}
	\label{fig:architecture}
\end{figure}


The contributions of this work are:
(i) a family of analysis tools that provide tighter guaranteed bounds on possible NN outputs for the same computational effort compared to the state-of-the-art,
(ii) two new partitioning algorithms that efficiently refine the input set partition based on desired output set shapes, \added{(iii) demonstrations of the proposed algorithms on NNs used in practice, including various NN architectures (e.g., fully connected, LSTM), deep NNs, and NNs with various nonlinear activations (e.g., ReLU, tanh)}, and (iv) applications on control systems to improve a robust RL agent's navigation efficiency by 22\% and reduce reachable set estimation error by 96\% for a robot arm.


\section{Background and Related Work}
\label{sec:background}

\textbf{Problem Statement:} 
Given a trained NN and a set of possible NN inputs, the objective is to find the tightest guaranteed over-approximation of the set of possible NN outputs.
The exact set of possible outputs is called the NN's \textit{reachable set}.
In general, finding the exact reachable set is computationally intractable for reasonably sized NNs; instead, the goal is to compute the over-estimate  $\mathcal{U}_e$ such that $\mathcal{U}\subseteq\mathcal{U}_e$ for a given input set.


\subsection{Propagators}
Propagators estimate how the full input set moves through the network, and they primarily differ in approximation strategies of the nonlinear activation functions.
At one extreme, Interval Bound Propagation (IBP)~\cite{gowal2018effectiveness} approximates the output of each layer with a tight $\ell_\infty$ ball, leading to conservative but fast-to-compute bounds of the final layer.
Convex relaxation-based techniques~\cite{salman2019convex} often achieve tighter bounds with more computation by approximating nonlinear activations with linear bounds -- some of these can be solved in closed-form~\cite{weng2018towards,zhang2018efficient}.
Other propagators provide tighter analysis at the cost of higher computation time, including approaches based on QP/SDP~\cite{Raghunathan_2018,fazlyab2019safety}, and convex relaxation refinements~\cite{singh2019beyond}.
While this paper focuses on analysis of trained NNs, several recent works consider the orthogonal problem of how to use these propagation techniques during training~\cite{zhang2019towards}.

\subsection{Partitioners}

\begin{figure}[t]
	\centering
	\includegraphics[page=5, height=0.25\textwidth, trim=0 0 0 0, clip]{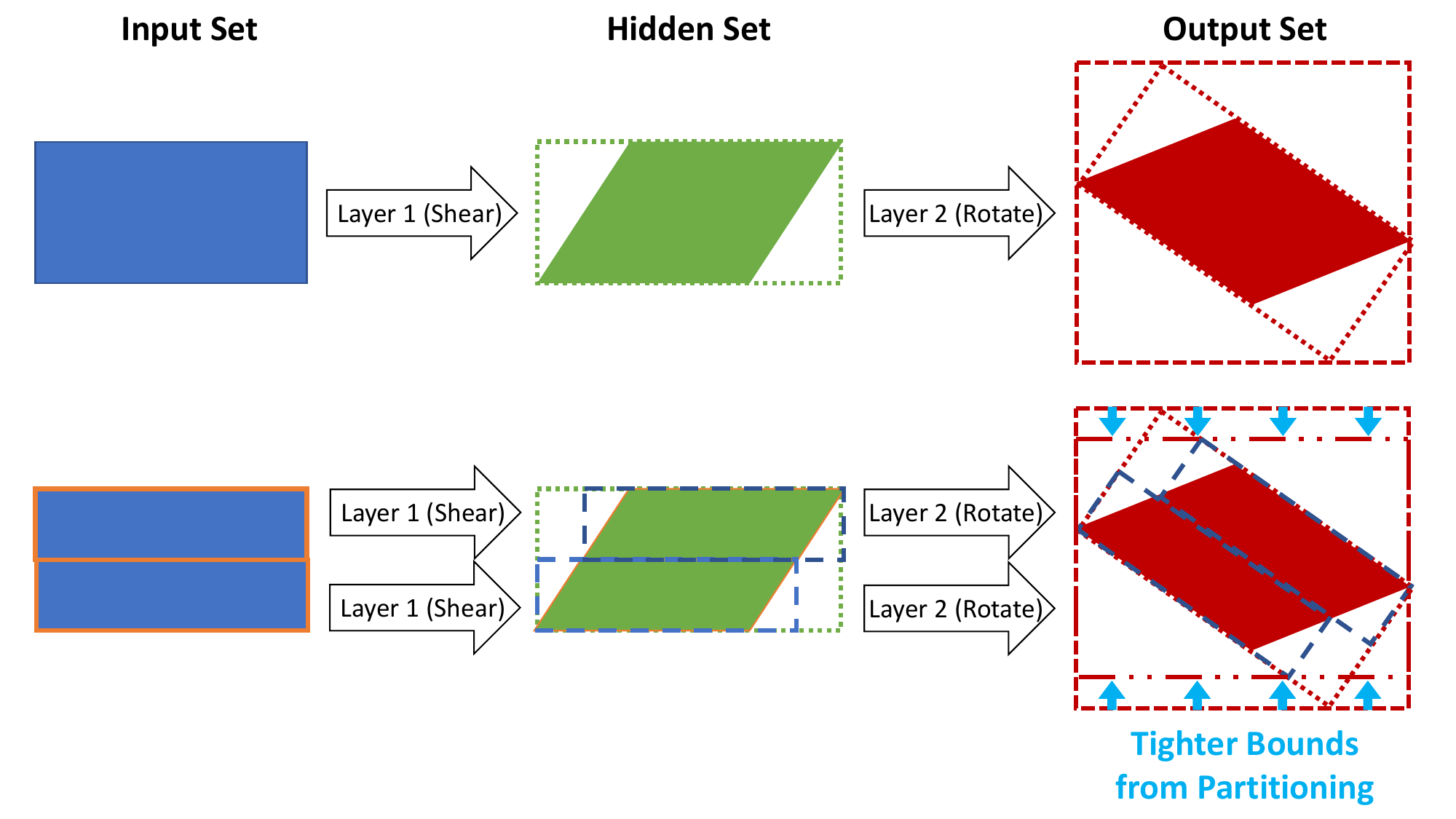}
	\caption{
	\small Partitioner Intuition.
	(Top) Large input sets cause loose bounds on NN output sets, even for this simple 2-layer NN with linear activations.
	(Bottom) Tighter bounds can be achieved by partitioning the input set, propagating each cell through the NN, and merging the output sets~\cite{xiang2018output,xiang2020reachable}.
	}
	\label{fig:partition_intuition}
\vspace{-0.2in}
\end{figure}

Partitioners break the input set into smaller regions, compute the reachable set of each small region, and return the total reachable set as the union of each smaller region's reachable set.
The idea is depicted in~\cref{fig:partition_intuition} for a simple NN with linear activations.
In the top row (without partitioning) IBP operates on the full input set, leading to excessive conservatism in the final output bound (top right: large red dashes vs. red shaded region).
The bottom row shows how IBP on two halves of the input set leads to a tighter approximation of the output set.

The key difference between partitioning approaches is the strategy for how to split the input set.
Some works make one bisection of the input set~\cite{anderson2020tightened}; \cite{xiang2018output} splits the input set into a uniform grid;
\added{\cite{wang2018formal} uses gradients to decide which cells to split for ReLU NNs}.
\added{
\cite{rubies2019fast} improves on~\cite{wang2018formal} using ``shadow prices'' to optimize \textit{how} to split a particular cell (i.e., along which dimension), but does not provide a way of choosing \textit{which} cells to split when computing tight reachable sets.
As illustrated in~\cite{xiang2020reachable}, substantial performance improvements can be achieved by stopping the refinement of cells that are already sufficiently refined.
Thus, the} current state-of-art partitioner, a Simulation-Guided approach (SG)~\cite{xiang2020reachable}, uses a partitioning strategy where Monte Carlo samples of the exact NN output are used as guidance for efficient partitioning of the input set, reducing the amount of computation required for the same level of bound tightness.
SG used IBP to compute output sets, and the two ideas of Partitioners and Propagators have been developed separately toward a similar objective.
This work addresses key gaps in the partitioning literature: we unify Partitioners with state-of-art Propagators for better performance, propose new partitioners that are flexible in the desired output set shape. 
We then show how improvements in robustness analysis map directly to reduced conservatism in control tasks.


\label{sec:approach}
\section{Approach}
This section introduces the overall architecture, describes our new partitioning algorithms, then analyzes the reduction in conservatism from partitioning.
\Cref{fig:architecture} shows a schematic of the proposed framework with its three nested modules: Analyzer, Partitioner and Propagator.
The Analyzer is aware of the desired output shape (e.g., lower bounds, $\ell_\infty$-ball, convex hull) and termination condition (e.g., computation time, number of Propagator calls, improvement per step).
The Analyzer specifies a Propagator (e.g., CROWN~\cite{zhang2018efficient}, IBP~\cite{gowal2018effectiveness}, SDP~\cite{fazlyab2019safety}, Fast-Lin~\cite{weng2018towards}) and a Partitioner (e.g., Uniform~\cite{xiang2018output}, Sim-Guided~\cite{xiang2020reachable} or the algorithms proposed in this section).
The Partitioner decides how to split the input set into cells, and the Propagator is used by the Partitioner to estimate the output set corresponding to an input set cell.

\subsection{Greedy Simulation-Guided Partitioning}
The state-of-art partitioning algorithm, SG~\cite{xiang2020reachable}, tightens IBP's approximated boundary with the following key steps:
(1) acquire $N$ Monte Carlo samples of the NN outputs to under-approximate the reachable set as the interval $[u_{\text{sim}}]$, (2) using IBP, compute the reachable set of the full input set and add this set to a stack $M$, and (3) (iteratively) pop an element from $M$, and either stop refining that cell if its computed reachable set is within $[u_\text{sim}]$, or bisect the cell, compute each bisection's reachable set, and add both to the queue.
The SG algorithm terminates when one of the cell's dimensions reaches some threshold, and the returned reachable set estimate is the weighted $\ell_\infty$-ball that surrounds the union of all of the cells remaining on the queue and $[u_{\text{sim}}]$.

\begin{figure}[t]
    \centering
    \begin{subfigure}{0.45\linewidth}

     \includegraphics[height=0.9\columnwidth]{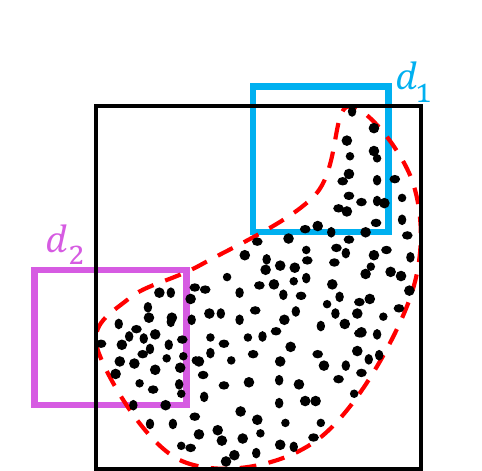}
 \caption{GSG Set Selection}
\label{fig:GSG_selection}
    \end{subfigure}
\begin{subfigure}{0.45\linewidth}
    \centering
    \includegraphics[height=0.9\columnwidth]{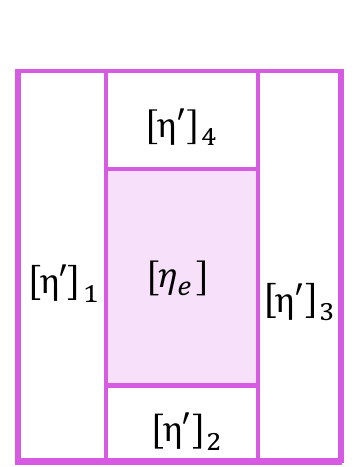}
    \caption{AGSG Decomposition}
    \label{fig:decompose_figure}
\end{subfigure}
    \caption{
    (a) GSG selects from candidates in $M$: the input set corresponding to the magenta output set is chosen for refinement, because $d_2>d_1$ (where $[u_{\text{sim}}]$ is the black box).
    (b) AGSG decomposition $[\eta]\setminus[\eta_e]$ into four new intervals $[\eta']_{1,2,3,4}$.
    }
    \label{fig:greedy_adaptive}
    \vspace{-0.1in}
\end{figure}

We propose a partitioning algorithm with better bound tightness for the same amount of computation, called Greedy-Sim-Guided (GSG), \added{described in Alg.~\ref{alg:AGSG},} by modifying the choice of which cell in $M$ to refine at each step.
Rather than popping the first element from the stack (LIFO) as in SG, GSG refines the input cell with corresponding output range that is furthest outside the output boundary of the $N$ samples (Line 17).
\added{This is illustrated in~\cref{fig:GSG_selection}, where the input cell corresponding to $d_2$ would be refined before $d_1$, because $d_2$'s output set (magenta) further exceeds the simulation-guided boundary estimate (black rectangle surrounding the black NN samples). 

Whereas SG might choose a cell that is not pushing the overall boundary outward at a given iteration, GSG will always choose to refine an input cell that is pushing the boundary.}
This heuristic gives the opportunity to reduce the boundary estimate at each iteration.
While the core SG algorithm remains the same, the greedy strategy can greatly improve the algorithm's performance.

\begin{algorithm}
\if \forarxiv 0
    \color{blue}
\fi
 \caption{\small Greedy Simulation-Guided Partitioning~\label{alg:AGSG}}
 \textbf{Input:} propagator operator $([.])$, termination condition ($\mathcal{T}_c()$), input interval ($[\eta]\subseteq \mathcal{H}$), Adaptive flag ($a =1$, if AGSG is active), Neural Network $\Phi$\\
 \textbf{Output:} output boundary   $\mathcal{U}_e$\\

\tcp{Initialization and MC sampling}
 $\mathbf{u}_{sim,n}= \Phi(x_{\textit{sim,n}}\in \mathcal{H}), n=1,\cdots,N$\\
 $[u_{\textit{sim}}] \gets \text{extrema of } u_{\textit{sim,n}}$ {\tcp{sim boundary}}
 $M \gets \emptyset, \mathcal{U}_e \gets \emptyset $\\
 \If{$a$}  
 {\tcp{AGSG: Adaptive Initialization}
 $(\eta^{*},{u}^{*}) \gets \arg\min\limits_{(\eta,u)}  \left\lVert u-\mu(\left[u_{\textit{sim}}])\right]\right\rVert_{2}$\\
 $\eta_{e} = \textit{Expand}(\eta^{*}$)\\ 
 $\{[\eta']_{i}\}=\textit{Decompose}(\{[\eta]\setminus[\eta_{e}]\}$) \tcp{\cref{fig:decompose_figure}}
 \For {$ [\eta']_{i} \in \{[\eta']_i\} $}
{
$[u_i]=[\Phi]([\eta]_{i})$\\
$M \gets M\cup\{([\eta]_{i},[u_{i}])\}\}$
 }
}
\Else
{\tcp{GSG: $M$ is full interval}
$[u]=[\Phi]([\eta])$\\
$M \gets M\cup \{([\eta],[u])\}$}
\tcp{Partition Refinement}
\While{$M\neq \emptyset$}{ 
               \tcp{\cref{fig:GSG_selection} (Pop from $M$)}
                $([\eta_w],[u_w]) \gets \argmax_{([\eta],[u])\in M} d([u], [u_\text{sim}])$
              
     \If{$[u_w] \in [\mathbf{u}_{\textit{sim}}]$}
            {$\mathcal{U}_e \gets \mathcal{U}_e \cup [u_w]$
           }
      \Else{
            \If {$\mathcal{T}_c()$}{break \tcp{Terminate}}
             \Else{
            $[\eta_1],[\eta_2]$ = Bisect($[\eta_w]$)\\
            $[u_1],[u_2] = [\Phi]([\eta_{i=1,2}])$\\
            $M \gets M \cup\{([\eta_1],[u_1])\} \cup\{ ([\eta_2],[u_2])\}$\\
          }
}
}

\Return $\mathcal{U}_e \gets \mathcal{U}_{e}\cup(\cup_{\{[\eta],[u]\}\in M^{[u]}})$
\end{algorithm}

\begin{algorithm}
\if \forarxiv 0
    \color{blue}
\fi
\caption{$Expand([\eta]^{(0)})$\label{alg:expand}}
\textbf{Input:} Initial input interval($[\eta]^{(0)}$),  step size($e_s$), NN propagator $[\Phi]$, simulated output set$[\mathbf{u}_{sim}]$\\
\textbf{Output:} expanded interval $[\eta_e]$\\
$[\eta_e] \gets [\eta]^{(0)} $\\
$[u] \gets [\Phi] ([\eta_{e}])$\\
\While {True}{
\If{$[u] \subset [u_{\textit{sim}}]$}
{
$[\eta_{e}]\gets[\eta_{e}]+e_s$ \\
$[u]= [\Phi]([\eta_e])$\\
}
\Else{$[\eta_e] \gets[\eta_e]-e_s$\\
\Return{$[\eta_e]$}
}
}
\end{algorithm}

\subsection{\added{Adaptive-Greedy-Simulation-Guided Partitioning}}

The Adaptive-Greedy-Sim-Guided (AGSG) algorithm extends GSG's initialization procedure to reduce wasted computation time refining unimportant input regions. When $a$ is activated, the AGSG initialization process is used (Lines 6-12).
After computing $[u_\text{sim}]$, AGSG initializes $[\eta_e]$ as the input point whose output is at the middle of simulated outputs.
This cell is repeatedly expanded with step size $e_s$ as long as its output set (computed by a Propagator) remains inside $[u_{\textit{sim}}]$. The \text{Expand} procedure is explained in Alg.~\ref{alg:expand}.
\added{The expanding cell is guaranteed to produce an output inside the simulated boundary.}
\begin{lemma}\label{lemma:adaptive}
\added{
 The overestimated output of the expanded interval $[\eta_e] \in \mathcal{H}$ to the neural network $\Phi$ when the output set is approximated by a propagator $[\Phi]$, is $[u_e]$ such that
\begin{align}
    [u_e]=[\Phi]([\eta_e]) \subseteq [u_{\textit{sim}}]
    \vspace{-20pt}
\end{align}
}
\end{lemma}
\begin{proof}
\added{
Assume the expanded interval and its output estimate at step $t$ are denoted by $[\eta_e]^{t}$ and $[u_e]^{t}$ respectively. If $[u_e]^{t}=[\Phi]([\eta_e]^{t}) \not \subset [u_{\textit{sim}}]$, there are two possible cases: (1) $[u_e]^{t-1} \subset [u_{\textit{sim}}]$ (the expansion of $[\eta_e]^{t-1}$ to $[\eta_e]^{t}$ causes this outcome). In this case, $[\eta_e]^{t}$ is reduced back to $[\eta_e]^{t-1}$ according to Alg.~\ref{alg:expand}'s expanding condition in line (9). Thus, $[\Phi]([\eta_e]^{t-1}) \subseteq [u_{\textit{sim}}]$ -- this condition is not stable. Case 2 would occur when $[u_e]^{t-1} \not \subset [u_{\textit{sim}}]$, but this is not possible, since the interval would never be expanded if  $[u_e]^{t-1} \not \subset [u_{\textit{sim}}]$, unless the initial interval's output is outside the simulated boundary: $[u_e]^{(0)} = [\Phi]([\eta_e]^{(0)})$. The initial interval's output cannot be outside the simulated boundary (contradiction), since $[\eta_e]^{(0)} $ is initialized via sampled inputs, thus $u^{(0)} \subset [u_{\textit{sim}}]$. Therefore the approximated output of the expanding input is always inside the simulated set.
}
\end{proof}
The remaining input is decomposed into a set of disjoint intervals $[\eta']_i$.
\cref{fig:greedy_adaptive} shows this decomposition in $2D$, which creates four new intervals $[\eta']_{1,2,3,4}$ (some of which could be empty) of rectangle shape\footnote{Extension of the decomposition to higher dims. is left as future work.}.
The new intervals $[\eta']_{i}$ are passed to GSG as the initial $M$.
The output set boundary estimate returned by AGSG merges that GSG output and the initial expanded cell's output boundary.






\subsection{Boundary Specification:} While SG only computes a $\ell_\infty$-ball over-approximation, GSG/AGSG optimize for the desired output set shape.
For example, if the goal is to find a convex hull over-approximation, GSG/AGSG modify the idea from~\cref{fig:greedy_adaptive} to select the input set that is furthest from the convex hull boundary (instead of the $\ell_\infty$-ball, as in SG~\cite{xiang2020reachable}).

\section{Experimental Results}\label{sec:result}
This section shows example partitions, applies the ideas to a robotic arm task, demonstrates better closed-loop behavior in collision avoidance, shows an ability to scale to various network sizes/architectures, and measures improvement along the time vs. tightness tradeoff.

\subsection{Partitions for Different Output Shapes}
\begin{figure}[t]
\centering
\begin{subfigure}{0.33\linewidth}
	\centering
	\includegraphics[width=\textwidth, trim=0 0 0 0, clip]{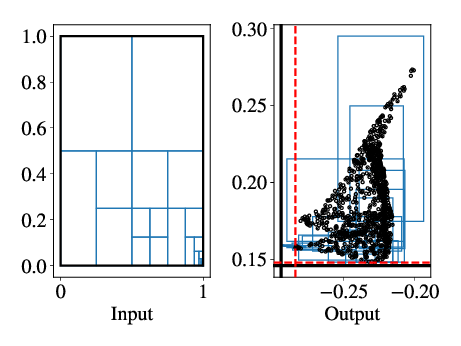}
	\caption{ Lower Bounds}
	\label{fig:tightest_lower_bounds}
\end{subfigure}%
\begin{subfigure}{0.33\linewidth}
	\centering
	\includegraphics[width=\textwidth, trim=0 0 0 0, clip]{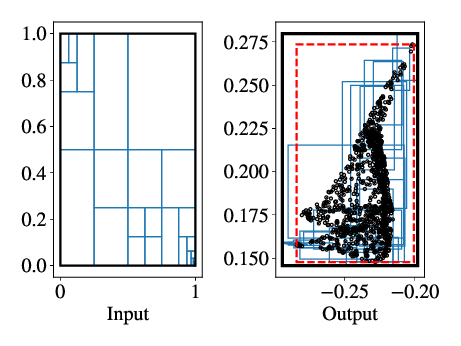}
	\caption{ $\ell_{\infty}$-ball}
	\label{fig:tightest_linf_ball}
\end{subfigure}%
\begin{subfigure}{0.33\linewidth}
	\centering
		\includegraphics[width=\textwidth, trim=0 0 0 0, clip]{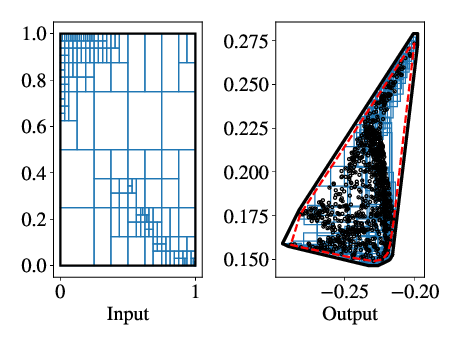}
	\caption{ Convex Hull}
	\label{fig:tightest_convex_hull}
\end{subfigure}%
\caption{\label{fig:boundary}
Input \& Output Sets for Different Output Set Shapes.
The estimated bounds (black) are ``tight'' when they are close to the bounds from exhaustive sampling (dashed red).
The GSG partitioner with CROWN~\cite{zhang2018efficient} propagator ran for 2 sec.
}
\label{fig:tightest_bounds_for_different_conditions}
\end{figure}

The ability to partition efficiently for different output shapes is shown in~\cref{fig:boundary} for a randomly initialized NN with 2 inputs, 2 outputs, and 50 nodes in hidden layer, \emph{i.e.,} $(2,50,2)$, with ReLU activations, and input set $[0,\,1]\times[0,\,1]$.
Each of (a-c) uses GSG with CROWN for 2 seconds.
Recall that SG~\cite{xiang2018output} would only return one output set for (a-c).

\begin{figure}
\centering
    \begin{subfigure}{0.25\textwidth}
        \centering
        \includegraphics[width = 0.8\textwidth, trim=40 20 40 20, clip]{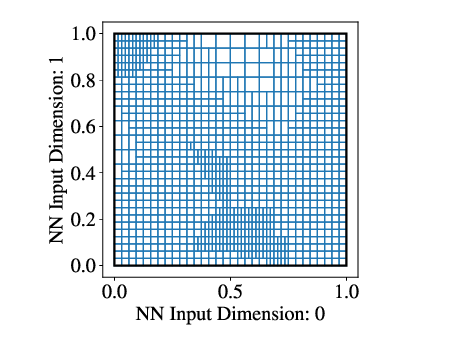}
        \captionsetup{justification=centering}
        \caption{ SG+IBP~\cite{xiang2020reachable}\\Partitions: 1021; Error: 0.35}
        \label{fig:SG+IBP_random}
    \end{subfigure}%
    \begin{subfigure}{0.26\textwidth}
        \centering
        \includegraphics[width=0.8\textwidth, trim =40 20 40 20, clip]{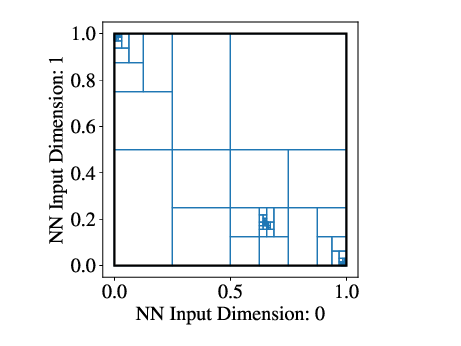}
        \captionsetup{justification=centering}
        \caption{SG+CROWN\\Partitions: 387; Error: 0.09}
        \label{fig:SG+CROWN_random}
    \end{subfigure}%
    \\
     \begin{subfigure}{0.26\textwidth}
        \centering
      \includegraphics[width=0.8\textwidth,trim=40 20 40 20, clip]{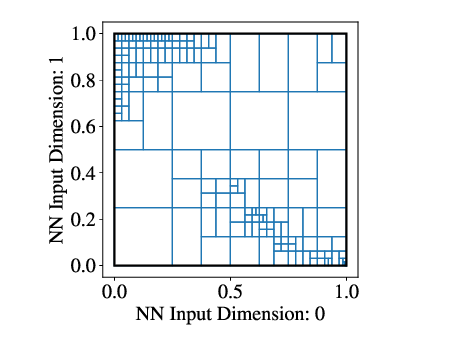}
        \captionsetup{justification=centering}
        \caption{GSG+CROWN\\Partitions: 280; Error: 0.02}
        \label{fig:GSG+CROWN_random}
    \end{subfigure}%
    \begin{subfigure}{0.26\textwidth}
        \centering
        \includegraphics[width=0.8\textwidth,,trim=40 20 40 20, clip]{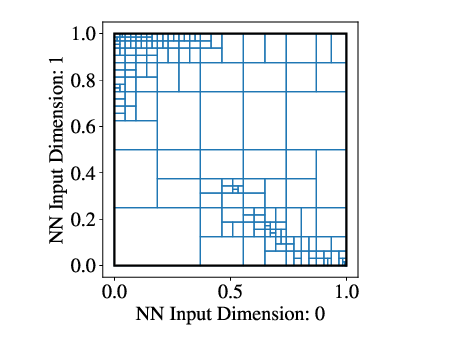}
        \captionsetup{justification=centering}
        \caption{AGSG+CROWN\\Partitions: 275; Error: 0.02}
        \label{fig:AGSG+CROWN_random}
    \end{subfigure}%
    \caption{Input partitions for a random (2, 50, 2) ReLU NN analyzed for 2 seconds. New methods (b, c, d) reduce the number of input partitions and output set (convex hull) error.}
    \label{fig:different_partition_approaches}
\end{figure}

\subsection{Comparisons to Baselines \& Ablation Study}
Four partitioning algorithms are compared in~\cref{fig:different_partition_approaches} for the same $(2,50,2)$ NN and input set.
Each analyzer runs for 2 seconds to compute an estimated output set.
The true output set is obtained by exhaustively sampling from the input space, and error is reported as percent extra area, $\frac{A_\text{estimate}-A_\text{true}}{A_\text{true}}$.
The proposed partitioning algorithms GSG (c) and AGSG (d) use only $280$ and $275$ partitions respectively, and their approximation error is $0.018$, which indicates more than $79\%$ improvement over SG -CROWN (b) and $95\%$ over the state of the art SG-IBP~\cite{xiang2020reachable}.
In addition to quantitative improvement, \cref{fig:different_partition_approaches} illustrates the input set partitions of each algorithm, which highlights how GSG refines different/fewer cells as SG, and that AGSG does not strictly make bisections.

\subsection{Applications in Robotics and Control}
\begin{figure}
\centering
    \begin{subfigure}{0.35\linewidth}
        \centering
    	\includegraphics[width=0.9\textwidth, trim =0 10 10 0 , clip]{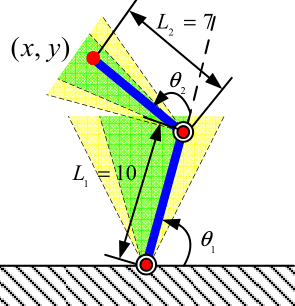}
    	\caption{ Robotic arm~\cite{xiang2018output}}
    	\label{fig:robot_arm} 
    \end{subfigure}%
    \begin{subfigure}{0.6\linewidth}
        \centering
        \tiny{
        \begin{tabular}{|c||c|c|c||}
            \hline
          \added{Algorithm} & \multicolumn{3}{c||}{Stats} \\
            \added{(Prop. + Part.)} & Error & Prop. Calls & Partitions \\ \hline
            IBP + SG~\cite{xiang2020reachable} & 0.216 & 1969 & 985  \\
            IBP + GSG & 0.042 & 869 & 435 \\
            IBP + AGSG & 0.040 & 847 & 425 \\
            \hline
            Fast-Lin + SG & 0.134 & 593 & 297 \\
            Fast-Lin + GSG & 0.009 & 473 & 237 \\
            Fast-Lin + AGSG & \textbf{0.008} & 461 & 232\\
            \hline
            CROWN + SG & 0.134 & 587 & 294 \\
            CROWN + GSG & 0.009 & 467 & 234 \\
            CROWN + AGSG & \textbf{0.008} & \textbf{453} & \textbf{228} \\
            \hline
        \end{tabular}
        }
        \caption{Comparison of Algorithms}
    \end{subfigure}\\
    \begin{subfigure}{0.5\linewidth}
        \centering
        \includegraphics[width =0.8\textwidth,  trim =40 20 50 20, clip]{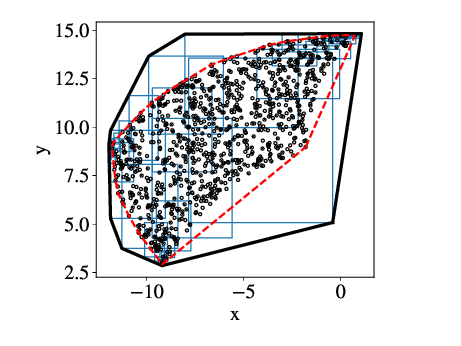}
        \caption{SG-IBP~\cite{xiang2020reachable}}
        \label{fig:SG+IBP_arm}
    \end{subfigure}%
    \begin{subfigure}{0.5\linewidth}
        \centering
        \includegraphics[width=0.8\textwidth,trim= 40 10 50 20,clip]{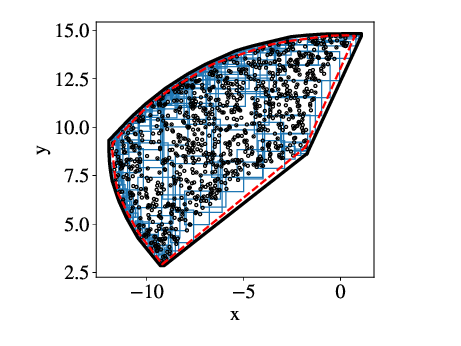}
        \caption{AGSG-CROWN}
        \label{fig:AGSG+CROWN_arm}
    \end{subfigure}%
    \caption{Reachable set estimate of a robotic arm. GSG/AGSG-CROWN achieves 96\% lower error than~\cite{xiang2020reachable} (2 sec.).}
\end{figure}

\subsubsection{Reachable Set Analysis of Robotic Arm}
Borrowing the robotic arm model from~\cite{xiang2018output,xiang2020reachable}, we compare our algorithm to~\cite{xiang2020reachable} for reachable set estimation of a forward kinematics model.
\cref{fig:robot_arm} shows the 2 DOF robot arm, with input $(\theta_1,\theta_2)$ as joint angles and output $(x,y)$ as end effector position.
The nonlinear dynamics are modeled by a small $(2,5,2)$ NN with tanh activations.
\added{One motivation for computing tight reachable sets here is to ensure that the robot arm will not collide with any obstacles, without being overly conservative.}
We assume a time limit of 2 sec to approximate the convex hull of end effector positions from the set of joint angles $(\theta_1,\,\theta_2)\in[\frac{\pi}{3},\,\frac{2\pi}{3}]\times[\frac{\pi}{3},\,\frac{2\pi}{3}]$.
As shown in~\cref{fig:robot_arm}, AGSG-CROWN reduces the error from~\cite{xiang2020reachable} by 96\%.
Only switching the partitioner (SG-IBP vs. (A)GSG-IBP) still achieves 80\% error reduction.
The estimated boundary is shown in (c, d).

\subsubsection{Multiagent Collision Avoidance}

Deep RL methods are popular in multiagent collision avoidance literature~\cite{long2018towards,chen2017decentralized}, but they rarely account for measurement uncertainty.
\cite{everett2020certified} proposed a certifiably robust deep RL algorithm, \added{which involves estimating a tight lower bound on the NN (e.g., DQN) output given that the agent could be within some state set.
In that work, large input uncertainties can degrade performance, partially due to overly conservative lower bounds from Fast-Lin~\cite{weng2018towards}.
This example motivates the need for tight reachable set estimation algorithms, as proposed in this work.}

\begin{figure}[!t]
  \begin{tabular}[b]{cc}
    \begin{tabular}[b]{c}
      \begin{subfigure}[b]{0.45\columnwidth}
      \centering
        \includegraphics[width=0.65\textwidth, trim=25 10 40 10, clip]{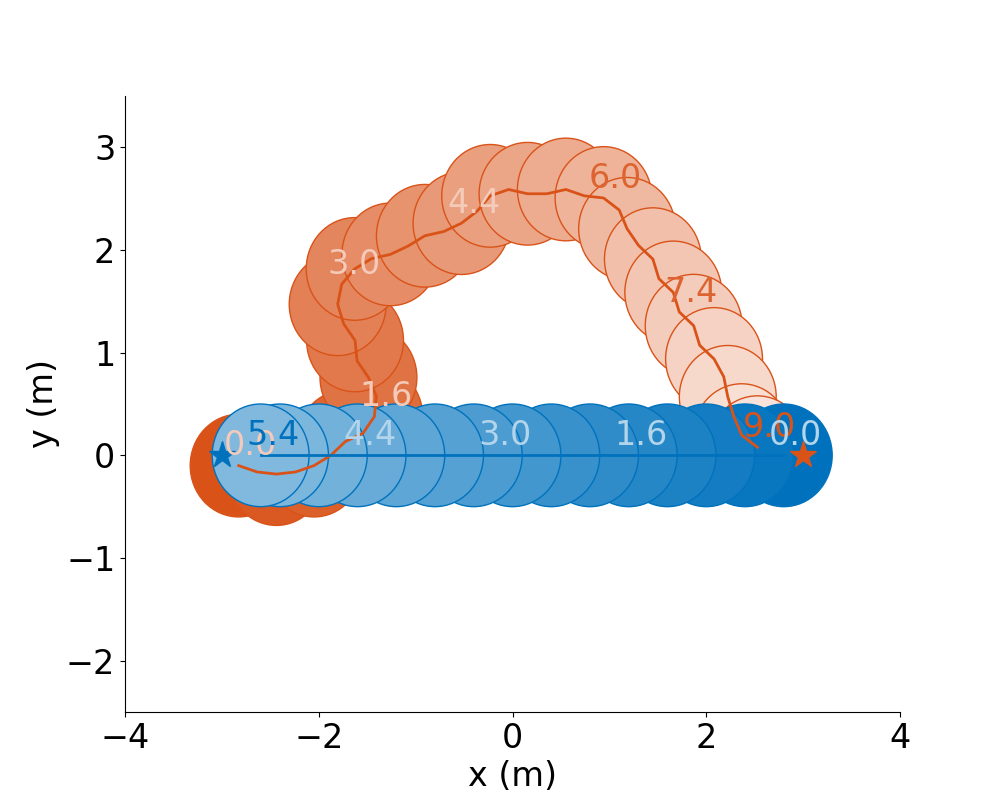}%
                        \captionsetup{justification=centering}
          \caption{\scriptsize CROWN (highly conservative)} 
        \label{fig:rl_crown}
      \end{subfigure}\\
      \begin{subfigure}[b]{0.45\columnwidth}
            \centering
        \includegraphics[width=0.65\textwidth, trim=45 10 40 10, clip]{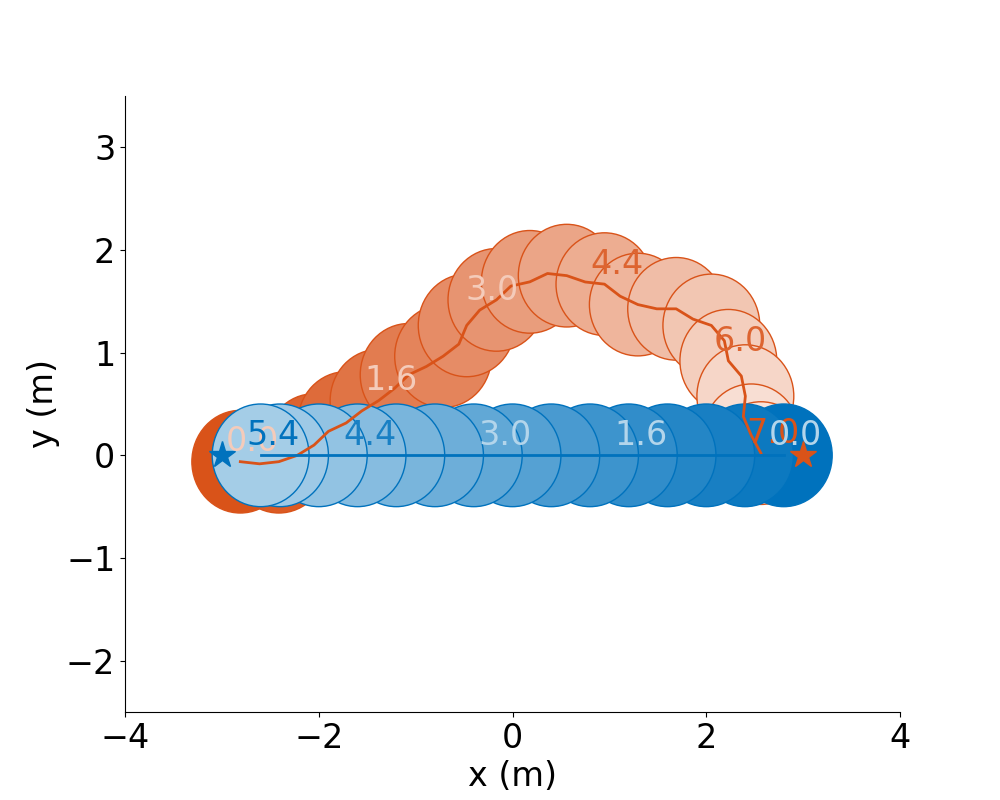}%
                \captionsetup{justification=centering}
        \caption{\scriptsize GSG-CROWN (less conservative)}
        \label{fig:rl_gsgcrown}
      \end{subfigure}
    \end{tabular}
    &
    \begin{subfigure}[b]{0.45\columnwidth}
        \includegraphics[width=\textwidth, trim=10 150 0 480, clip]{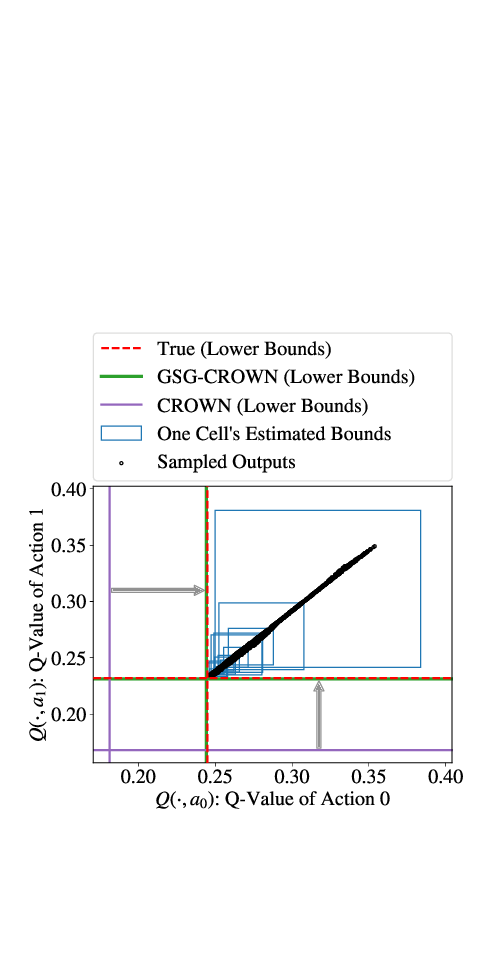}
        \captionsetup{justification=centering}
        \caption{\scriptsize Worst-Case Q-Values (Model-Free RL)}
        \label{fig:rl_bounds}
    \end{subfigure}
  \end{tabular}
\label{fig:applications}
\vspace{-0.1in}
  \caption{
Multiagent collision avoidance under uncertainty.
In (a), a robust but conservative trajectory from a robust RL formulation~\cite{everett2020certified} that used CROWN/Fast-Lin~\cite{zhang2018efficient} to estimate worst-case Q-values under uncertainty on obstacle positions.
In (b), the proposed GSG-CROWN algorithm enables the orange agent to reach its goal faster (7 vs. 9 sec) while still avoiding the blue agent.
This improved behavior is a result of tighter estimates of worst-case Q-values, shown at a single timestep in (c) (purple $\to$ green).}
\vspace{-0.3in}
\end{figure}

The robust-but-conservative behavior caused by CROWN/Fast-Lin is shown in~\cref{fig:rl_crown} ($\pm0.5$m uncertainty on the blue agent's position at each timestep).
By instead using GSG-CROWN to estimate worst-case Q-values (\cref{fig:rl_gsgcrown}), the orange agent reaches the goal much faster while still avoiding the blue agent.
This improved behavior is a result of tighter estimates of worst-case Q-values, shown at a single timestep in \cref{fig:rl_bounds}.
For this experiment, a (11, 64, 64, 11) DQN (11 states \& 11 discrete actions) was trained with perfect measurements in the \texttt{gym-collision-avoidance} environment~\cite{Everett18_IROS}.
Furthermore, this application is a case where only \textit{lower bounds} on the NN outputs are needed, which motivates the use of our proposed algorithms that can focus computation toward this objective.

\subsection{Scalability to Larger NNs}
The proposed algorithms scale to bigger NNs as shown in~\cref{tab:scalability} and improve on existing methods by orders of magnitude.
For a given number of propagator calls (100), the average error (close to 0 is best) of 10 random NNs is reported for 4 different architectures (small, deep, higher dimension, LSTM), and for different boundary types.
The (4, 100, 10) NN has uncertainty on the full 4D input set and the LSTM uses 2D uncertainty on the last timestep.
While this work's approaches scale well to deep NNs and various architectures, future work should consider the challenges from settings with even higher dimensional input uncertainties.

\begin{table}[t]

\tiny
\begin{tabular}{|m{0.3cm}|c||c|c|c||}
\hline
NN & \added{Algorithm} & \multicolumn{3}{c||}{Boundary Type} \\
& \added{(Prop. + Part.)} & Lower Bounds & $\ell_\infty$-ball & Convex Hull \\ \hline
\multirow{10}{*}{\STAB{\rotatebox[origin=c]{90}{(2, 100, 2)}}} 
 & IBP~\cite{gowal2018effectiveness} & \num{1.50e+00} & \num{7.77e+01} & \num{9.06e+00} \\
& IBP + SG~\cite{xiang2020reachable} & \num{2.47e-01} & \num{4.06e+00} & \num{1.49e+00} \\
& IBP + GSG & \num{1.70e-01} & \num{3.44e+00} & \num{1.44e+00} \\

& Fast-Lin~\cite{Weng_2018} & \num{2.78e-01} & \num{4.62e+00} & \num{1.90e+00} \\
& Fast-Lin + SG & \num{1.70e-03} & \num{1.70e-02} & \num{1.12e-01} \\
& Fast-Lin + GSG & \num{3.94e-03} & \num{5.48e-02} & \num{7.23e-02} \\


& CROWN~\cite{zhang2018efficient} & \num{2.15e-01} & \num{3.29e+00} & \num{1.55e+00} \\
& CROWN + SG & \textbf{\num{1.34e-03}} & \textbf{\num{1.23e-02}} & \num{1.09e-01} \\
& CROWN + GSG & \num{3.49e-03} & \num{5.32e-02} & \textbf{\num{6.65e-02}} \\
& SDP~\cite{fazlyab2019safety} & \num{1.20e-01} & \num{1.90e+00} & \num{1.06e+00} \\
\hline
\hline
\multirow{9}{*}{\STAB{\rotatebox[origin=c]{90}{\parbox{2cm}{\centering (2, 100, 100, 100, 100, 100, 100, 2)}}}} 
& IBP~\cite{gowal2018effectiveness} & \num{1.69e+02} & \num{8.17e+09} & \num{1.07e+05} \\
 & IBP + SG~\cite{xiang2020reachable} & \num{3.32e+01} & \num{3.16e+08} & \num{2.10e+04} \\
 & IBP + GSG & \num{3.07e+01} & \num{2.67e+08} & \num{1.93e+04} \\
& Fast-Lin~\cite{Weng_2018} & \num{2.32e+00} & \num{1.57e+06} & \num{1.48e+03} \\
& Fast-Lin + SG & \num{2.65e-04} & \num{3.83e-01} & \num{2.91e-01} \\
& Fast-Lin + GSG & \num{9.32e-05} & \num{2.12e-01} & \num{2.30e-01} \\

& CROWN~\cite{zhang2018efficient} & \num{8.96e-01} & \num{2.42e+05} & \num{5.74e+02} \\
& CROWN + SG & \num{1.61e-04} & \num{2.03e-01} & \num{1.91e-01} \\
& CROWN + GSG & \textbf{\num{5.40e-05}} & \textbf{\num{1.18e-01}} & \textbf{\num{1.65e-01}} \\

\hline
\hline
\multirow{9}{*}{\STAB{\rotatebox[origin=c]{90}{(4, 100, 10)}}} & 
IBP~\cite{gowal2018effectiveness} & \num{3.11e+01} & \num{1.41e+17}&- \\
& IBP + SG~\cite{xiang2020reachable} & \num{1.40e+01} & \num{6.89e+13}&- \\
&IBP + GSG & \num{1.33e+01} & \num{4.37e+13}&- \\
& Fast-Lin~\cite{Weng_2018} & \num{6.18e+00} & \num{5.95e+10}&- \\
& Fast-Lin + SG & \num{8.24e-01} & \num{7.81e+03}&- \\
& Fast-Lin + GSG & \num{7.44e-01} & \num{4.59e+03}&- \\
& CROWN~\cite{zhang2018efficient} & \num{4.51e+00} & \num{4.01e+09} & -\\
&CROWN + SG & \num{5.60e-01} & \num{8.17e+02} &-\\
& CROWN + GSG & \textbf{\num{5.00e-01}} & \textbf{\num{5.52e+02}} &-\\
\hline\hline

\multirow{3}{*}{\STAB{\rotatebox[origin=c]{90}{\parbox{0.8cm}{\centering \vspace{-0.05in}LSTM\\((8, 8), \\64, 2)}}}}& IBP~\cite{gowal2018effectiveness} & \num{1.56e-02} & \num{1.16e+02} & \num{1.13e+01} \\
& IBP + SG~\cite{xiang2020reachable} & \num{2.44e-03} & \num{5.55e+00} & \num{1.83e+00} \\
& IBP + GSG & \textbf{\num{1.80e-03}} & \textbf{\num{4.90e+00}} & \textbf{\num{1.80e+00}} \\
\hline
\end{tabular}

\caption{
Approximation Error (closer to 0 is better) for four different types of NNs: small, deep, higher dimension, and LSTM.
Reported values are average error across 10 randomly initialized NNs after up to 100 propagator calls.
}
\label{tab:scalability}
\vspace{-0.2in}
\end{table}

\subsection{Computation-Performance Tradeoff}
\begin{figure}
    \centering
    \begin{subfigure}{0.48\linewidth}
    \centering
     \includegraphics[width =0.9\textwidth,height =0.7\textwidth, trim=10 20 20 20,clip]{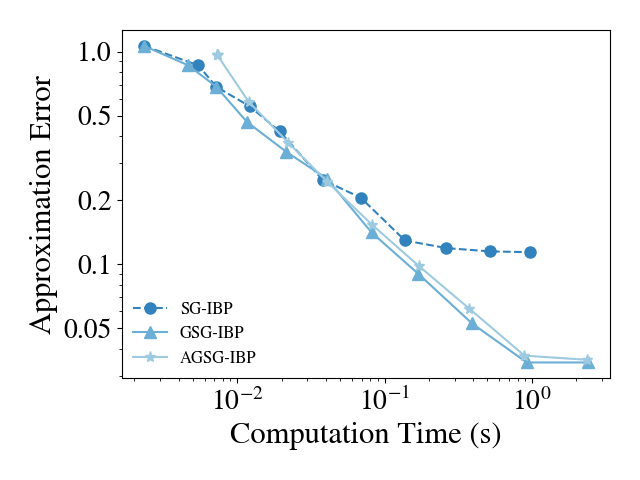}
    \caption{\footnotesize{Propagator: IBP}}
    \label{fig:ibp}       
    \end{subfigure}
    \begin{subfigure}{0.48\linewidth}
     \includegraphics[width =\textwidth,height =0.7\textwidth,trim=10 20 20 20, clip]{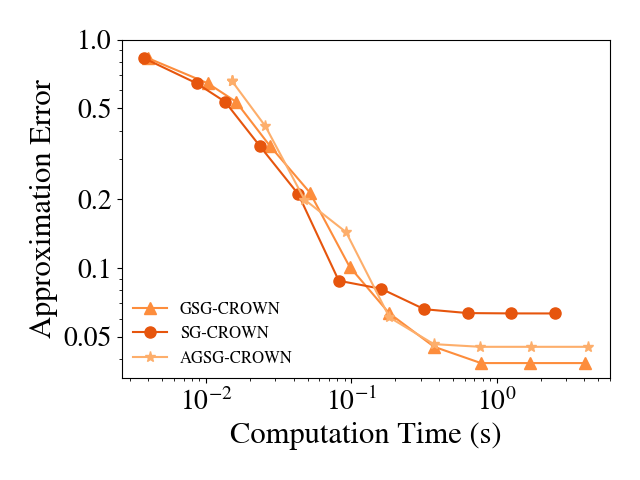}
    \caption{\footnotesize{Propagator: CROWN}}
    \label{fig:crown}       
    \end{subfigure}
    \begin{subfigure}{0.49\linewidth}
    \includegraphics[width =\textwidth,height =0.7\textwidth,trim=10 20 20 20, clip]{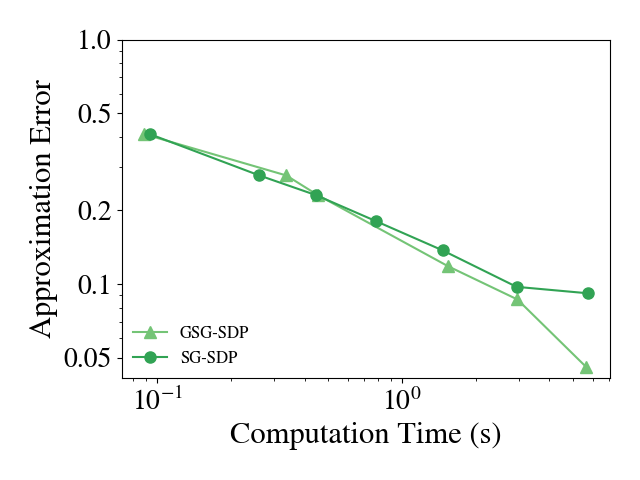}
    \caption{\footnotesize{Propagator: SDP}}
    \label{fig:sdp}       
    \end{subfigure}
        \begin{subfigure}{0.49\linewidth}
            	\centering
    	\tikzstyle{line} = [draw, -latex']
    	\begin{tikzpicture}
        	\node[anchor=south west,inner sep=0] (image) at (0,0)     
    { \includegraphics[width =\textwidth,height =0.75\textwidth,trim=10 20 20 20, clip]{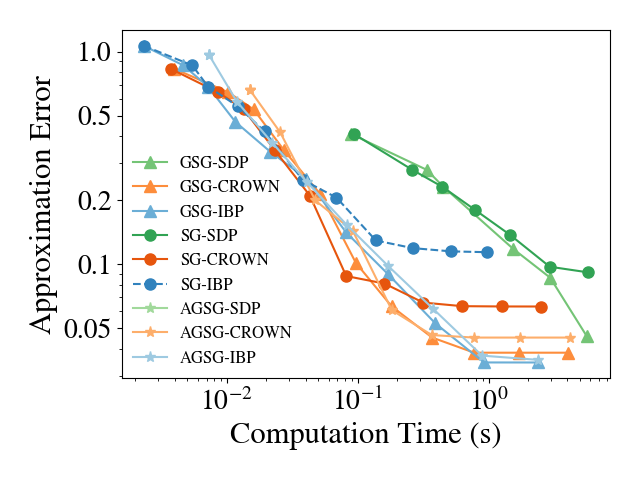}};
            \node [anchor=west] (crown) at (1.5,3) {\tiny{CROWN~\cite{zhang2018efficient}}};
            \node (crown_pt) at (1.1, 2.85){};
             \node (crown_start_pt) at (1.7,3.1) {};
            \draw[red, thick] (crown_pt) circle (0.1cm);
            \draw[red, thick,->] (crown_start_pt) edge (crown_pt);
            
            \node [anchor=west] (sdp) at (2.5,2.8) {\tiny{ SDP~\cite{fazlyab2019safety}}};
             \node (sdp_pt) at (2.4,2.4){};
             \node (sdp_start_pt) at (2.7,2.9) {};
             \draw[red,  thick] (sdp_pt) circle (0.15cm);            \draw[red,thick,->] (sdp_start_pt) edge (sdp_pt);
            
            \node [anchor=west,align=center] (sg) at (3., 2.3) {\tiny{SG-IBP~\cite{xiang2020reachable}}};
            \node [anchor=west,align=center] (sg_description) at (3., 2.1){\tiny{(Dashed Blue)}};
         \node (sg_pt) at (2.8,1.55){};
         \node (sg_start_pt) at (3.4, 2.1){};
           \draw[red,  thick,->]  (sg_start_pt) edge (sg_pt);

    
        \end{tikzpicture}%
        \caption{All Propagators (overlayed)}
    \label{fig:time_all}       
    \end{subfigure}
    \caption{
\added{Improvement in Computation Time vs. Accuracy Tradeoff.
(a-c) each compare different partitioners for a single propagator, (d) overlays (a-c) in one figure.
Colors indicate Propagator; markers indicate Partitioner.
The GSG/AGSG approaches outperform the SG approaches as the bounds are refined.
This work unified partition and propagation ideas to give many new methods (everything without a red annotation) that exceed the state-of-the-art.}}
\label{fig:num_splits_vs_bounds}
\end{figure}
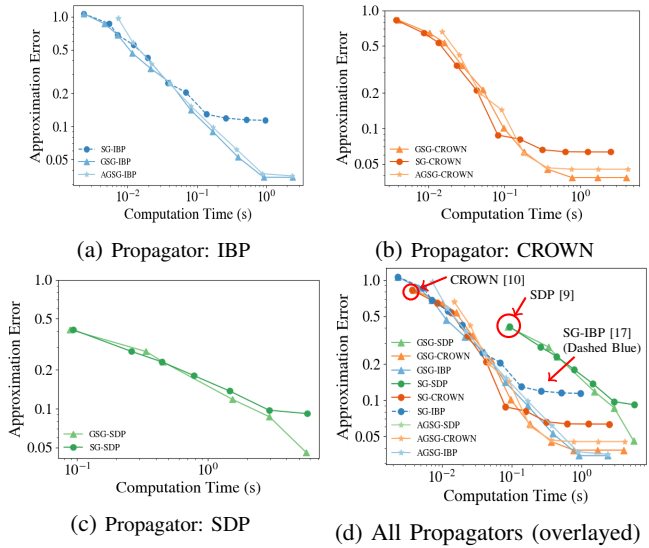
Throughout this paper, we have leveraged the idea of partitioning the input set to tighten the approximated boundary.
To empirically show that bounds tighten with additional computational effort, we plot several combinations of partitioners and propagators in~\cref{fig:num_splits_vs_bounds} over time.
Each color corresponds to a propagator (IBP, CROWN, SDP) and each marker corresponds to a partitioner (SG, GSG, AGSG).
This result uses the robotic arm model \added{and convex hull boundaries} from before, but with ReLU activations.

A first key takeaway is that additional computation time leads to reduced error (increased tightness)
Another key takeaway is that our framework provides many algorithms that exceed the performance of previous state-of-art algorithms~\cite{xiang2020reachable,zhang2018efficient,fazlyab2019safety}.
Except the blue dashed line~\cite{xiang2020reachable} and leftmost green/orange points~\cite{fazlyab2019safety,zhang2018efficient}, all of the options are new algorithms proposed by this work.
The analysis provided in the plots informs the choice of propagator and partitioner for a particular application with, say, a desired level of accuracy or budgeted resources (memory/computation).
Overall, for this task GSG-CROWN almost always provides the best accuracy vs. computation time tradeoff, requiring $\sim5\times$ less computation for the same accuracy as SG-IBP~\cite{xiang2020reachable}.


\section{Conclusion}
This work proposed a suite of algorithms for online robustness analysis of NNs that can provide confidence in NN decisions under uncertainty.
We build on recent work for handling large uncertainties by proposing new, flexible partitioning algorithms and give theoretical rationale for partitioning as a strategy for reducing conservatism.
Furthermore, we show how recent methods that efficiently relax NN nonlinearities can be unified with partitioning in a single framework, which provides many new state-of-art algorithmic choices for robotics applications.
Along with showing improved aggregate performance on random NNs with various sizes/architectures, we show how these ideas can be applied to other learning tasks for control systems, showing a 22\% improvement in robust RL for multiagent collision avoidance and a 96\% reduction in conservatism for a learned robotic arm kinematic model.



\bibliographystyle{ieeetr} 
\tiny
\bibliography{biblio}
\end{document}